\documentclass{article}
\usepackage{ijcai25}

\usepackage{times}
\usepackage{soul}
\usepackage{url}
\usepackage[hidelinks]{hyperref}
\usepackage[utf8]{inputenc}
\usepackage[small]{caption}
\usepackage{graphicx}
\usepackage{amsmath}
\usepackage{amsthm}
\usepackage{booktabs}
\usepackage{algorithm}
\usepackage{algorithmic}
\usepackage[switch]{lineno}
\usepackage{multirow}

\title{Causal-aware Large Language Models: Enhancing Decision-Making Through Learning, Adapting and Acting}

\author{
Wei Chen$^{1}$\and
Jiahao Zhang$^{1}$\and
Haipeng Zhu$^{1}$\and
Boyan Xu$^{1}$\and
Zhifeng Hao$^{1,2}$\and
Keli Zhang$^{3}$\and
Junjian Ye$^{3}$\and
Ruichu Cai$^{1,4}$ \footnote{Corresponding Author}  \\
\affiliations
$^1$ School of Computer Science, Guangdong University of Technology, Guangzhou, China \\
$^2$ College of Mathematics and Computer Science, Shantou University, Shantou, China\\
$^3$ Huawei Noah’s Ark Lab, Huawei, Paris, France\\
$^4$ Peng Cheng Laboratory, Shenzhen, China
\emails
\{chenweidelight,luckyanjooo,zhuhaipeng0514,hpakyim\}@gmail.com, haozhifeng@stu.edu.cn, \{zhangkeli1, 
yejunjian\}@huawei.com,cairuichu@gmail.com
}

\begin{document}

\maketitle

\begin{abstract}
Large language models (LLMs) have shown great potential in decision-making due to the vast amount of knowledge stored within the models.
However, these pre-trained models are prone to lack reasoning abilities and are difficult to adapt to new environments, further hindering their application to complex real-world tasks. To address these challenges, inspired by the human cognitive process, we propose Causal-aware LLMs, which integrate the structural causal model (SCM) into the decision-making process to model, update, and utilize structured knowledge of the environment in a ``learning-adapting-acting" paradigm.
Specifically, in the learning stage, we first utilize an LLM to extract the environment-specific causal entities and their causal relations to initialize a structured causal model of the environment. 
Subsequently, in the adapting stage, we update the structured causal model through external feedback about the environment, via an idea of causal intervention. 
Finally, in the acting stage, Causal-aware LLMs exploit structured causal knowledge for more efficient policy-making through the reinforcement learning agent. 
The above processes are performed iteratively to learn causal knowledge, ultimately enabling the causal-aware LLMs to achieve a more accurate understanding of the environment and make more efficient decisions. Experimental results across 22 diverse tasks within the open-world game ``Crafter" validate the effectiveness of our proposed method.
\end{abstract}
\begin{figure}[t]
    \centering
    \includegraphics[width=\linewidth]{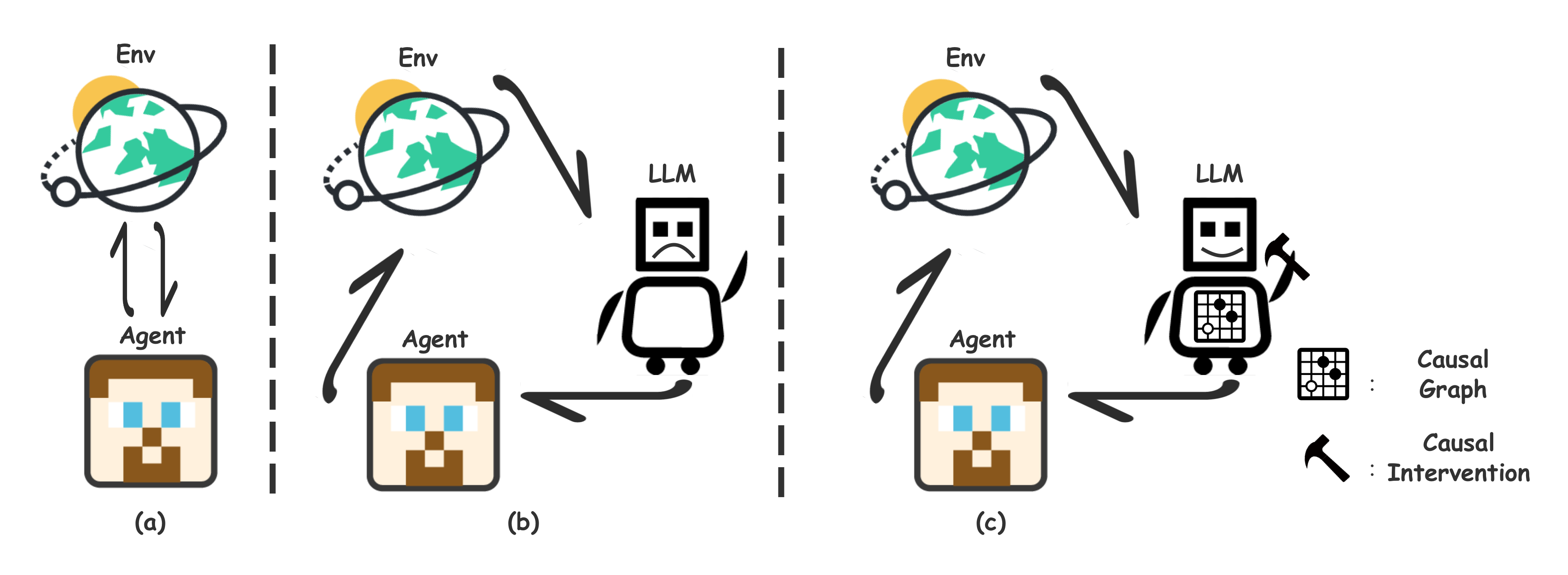}
    \caption{Comparison of Existing Approaches and Our Method: Previous works use (a) \textbf{RL agents alone} or (b) \textbf{LLMs to assist RL agents} in interacting with the environment. In (c) our \textbf{Causal-aware LLMs} framework, we incorporate causal knowledge about the environment to enhance the LLM's ability to understand it, thereby supporting RL agents in more effectively interacting with the environment and improving policy learning.}
    \label{fig: intro}
\end{figure}
\section{Introduction}
With the advancement of large language models (LLMs) like Generative Pre-trained Transformer (GPT) \cite{brown2020language} and Large Language Model Meta AI‌ (LLAMA) \cite{touvron2023llama}, these models have demonstrated significant potential in the realm of decision-making \cite{sun2023adaplanner,yao2023react}. 
However, the primary focus of these pre-trained models is to predict the next token based on the existing data, limiting their ability to reason structurally and adapt to new environments. These limitations hinder their effectiveness in real-world, complex decision-making tasks. As a result, effectively leveraging the capabilities of LLMs to solve complex tasks remains an ongoing research problem. 

To enhance LLMs' reasoning abilities and environmental adaptability, one promising approach is to incorporate reinforcement learning (RL) techniques. As shown in Figure \ref{fig: intro}(a), the \textbf{RL agents alone} methods \cite{kiran2021deep,kober2013reinforcement,shao2019survey} involve an agent interacting with its environment by selecting actions based on its current state, receiving rewards or penalties, and adjusting its strategy with the goal of maximizing long-term rewards. To improve an agent's ability to extract and interpret environmental data, the methods using \textbf{LLMs to assist RL agents} \cite{du2023guiding,baumlivision,zhang2024adarefiner}, have explored integrating LLMs to help the agent better understand the context of its environment, thereby enhancing its policy learning (see Figure \ref{fig: intro}(b)). Despite this advancement, these approaches still rely on the next-token prediction paradigm, which limits the agent's capacity for structured reasoning and rapid adaptation to unfamiliar or complex environments. This gap highlights the need for more advanced inherent knowledge management mechanism that go beyond token prediction and facilitate deeper reasoning and flexibility in decision-making processes.

The human cognitive processes (e.g. System 1 \& 2 ) offer valuable insights into addressing the aforementioned limitations \cite{kahneman2011thinking}. Intuitively, when interacting with a new environment, System 1 (intuitive, fast) and System 2 (analytical, slow) work in concert to acquire and process knowledge effectively. As humans adjust to the environment through actions, System 2 carefully evaluates feedback and refines strategies, while System 1 handles increasingly familiar aspects. Through iterative interactions, humans continuously refine their strategies based on environmental feedback, gradually transforming complex analytical processes into intuitive responses. This cognitive architecture suggests the necessity of a knowledge representation mechanism that supports both rapid retrieval and deliberate reasoning. Specifically, two key challenges arise: 1) how to represent and learn knowledge to understand the environment, and 2) how to adaptive update knowledge and reason to solve complex tasks in dynamic environments.

Based on the above analysis, we find that Structural Causal Models (SCMs) offer a robust framework for knowledge presentation and further provide a potential mechanism for reasoning and adaptive learning. To this end, we introduce a causal-aware LLMs framework (illustrated in Figure \ref{fig: intro}(c)), which integrates SCMs into LLMs to improve their ability to understand environments, acquire new knowledge, and adapt through dynamic interactions. The framework consists of three key stages: \textbf{learning}, \textbf{adapting}, and \textbf{acting}. In the \textbf{learning} stage, the LLM extracts causal knowledge from the environment to build an initial causal representation. In the \textbf{adapting} stage, we address potential inaccuracies in the causal relationships—often arising from the LLM’s tendency to generate hallucinated information—by applying causal intervention techniques to refine the causal graph iteratively. This intervention ensures the model better captures the true structure of the environment. Finally, in the \textbf{acting} stage, the updated causal relationships are integrated into both the LLM and the RL agent, enhancing the agent’s understanding of the environment and improving its policy learning. This iterative cycle allows the LLM and RL agent to collaboratively refine their knowledge and strategies, optimizing decision-making until the desired rewards are achieved. This approach promotes more robust and adaptive decision-making, enabling the model to tackle complex, real-world tasks effectively. Experimental results across 22 diverse tasks in the open-world game ``Crafter" demonstrate the efficacy of our proposed method in improving decision-making.
\begin{figure*}[ht]
    \centering
    \includegraphics[scale=0.1]{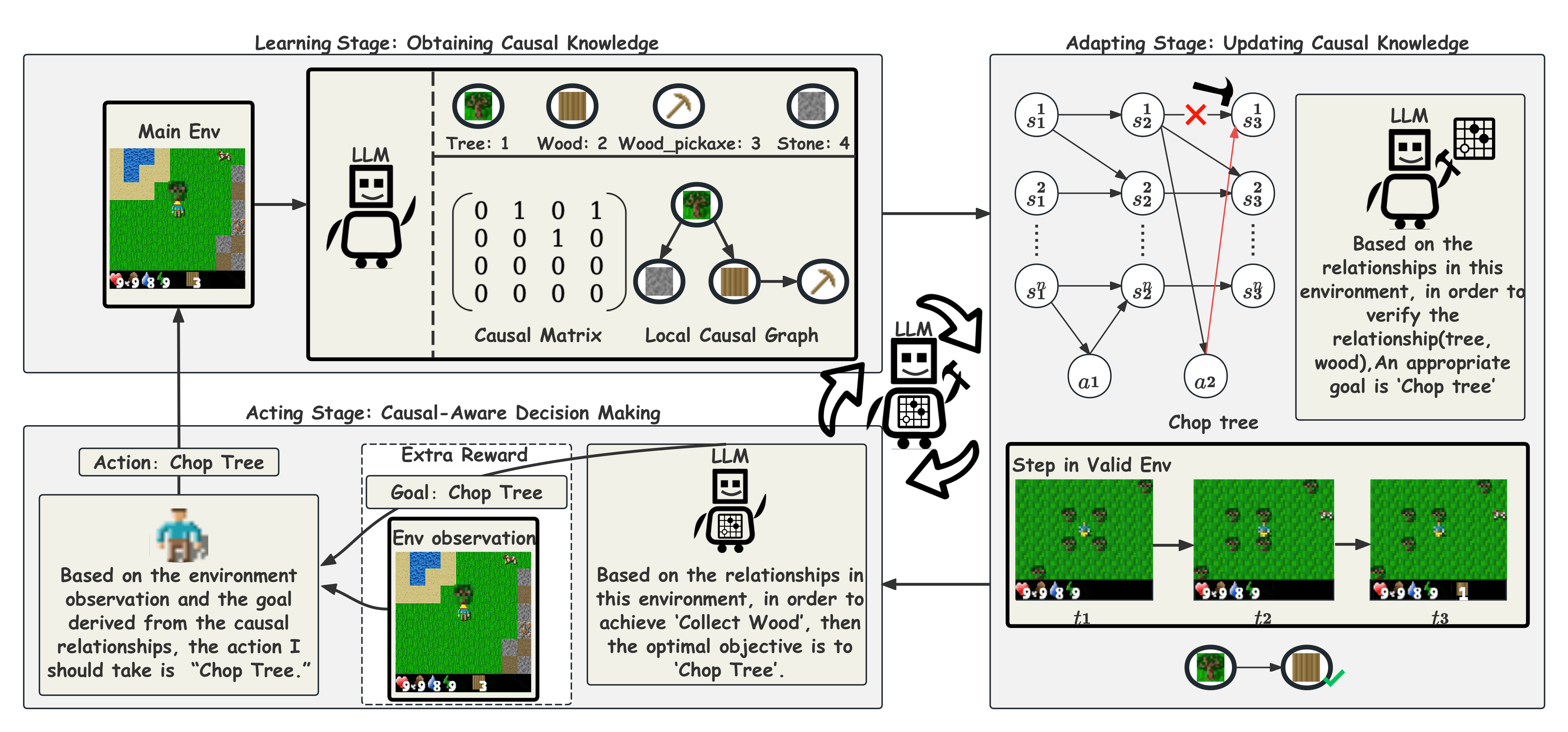}
    \caption{Our framework consists of three stages. In the \textbf{learning} stage, we use LLMs to learn causal knowledge among variables from the data collected from the environment. In the \textbf{adapting} stage, we update the causal knowledge by causal intervention, where $s^i_t$ represents the state of object $i$ at time $t$ and $a_i$ denotes the action performed by an agent. In the \textbf{acting} stage, we use LLM to assist the RL agent in policy learning by acting based on the learned causal knowledge for decision-making.}
    \label{framework}
\end{figure*}
\section{Related Work}
\subsection{LLM for Causality} Numerous scholars have conducted extensive research on whether large language models are capable of understanding causal relationships, which can be divided into exploring LLMs’ understanding of causal knowledge and using them to construct causal graphs. Research such as \cite{gao-etal-2023-chatgpt,ashwani2024cause,chen-etal-2024-clear} have shown that LLMs can handle some basic tasks related to causal graphs, but they struggle with more complex and nuanced tasks. While LLMs can mimic causal language, they need more explicit knowledge and reasoning mechanisms to truly understand causal relationships. 

\cite{zhang2024causal} use LLMs to recover causal graphs by retrieving relevant text, using LLMs to identify factor associations, and aggregating these relationships. But their effectiveness is limited without a causal discovery algorithm. Hallucination also limits LLMs’ ability to handle causal knowledge due to the lack of mechanisms to validate and constrain their generated causal relationships. We propose a unique technique that leverage causal intervention to refine and validate the causal relationships generated by LLMs, ensuring they align with environment-specific causal structures and improving their utility in downstream applications.
\subsection{LLM for Reinforcement Learning} 
Recent studies have investigated the integration of LLMs into RL, particularly for reward generation and the use of fine-tuning LLMs for policy optimization. These methods leverage the power of pre-trained LLMs to guide RL agents in complex tasks.
For reward generation, \cite{du2023guiding} proposed a framework called ELLM, which generates potential goals relevant to the agent’s current state using LLMs, offering a structured approach to reward shaping. Similarly, \cite{baumlivision} explored the use of Vision-Language Models (e.g., CLIP) to derive rewards for RL agents targeting language-specified visual goals. 

For fine-tuning LLMs, \cite{zhang2024adarefiner} introduced AdaRefiner, which utilizes a fine-tuned LLM as an adapter to refine task understanding based on agent feedback, enhancing the collaboration between LLMs and RL agents.
Although existing methods have demonstrated a certain degree of effectiveness, whether LLMs can adapt to diverse environments remains a pressing challenge. To address this issue, we propose a framework that integrates SCM to enhance the adaptability of LLMs in complex and dynamic environments.

\subsection{Reinforcement Learning} 
Reinforcement learning aims to learn an optimal policy to maximize the expected cumulative reward, fundamentally optimizing the decision-making process through trial and error. Common approaches include value-based methods and policy-based methods. Value-based methods such as Q-learning \cite{watkins1989learning}, DQN \cite{mnih2013playing}, and DDQN \cite{lillicrap2015continuous} find the optimal policy by estimating the value of states or state-action pairs. Policy-based methods such as PPO \cite{schulman2017proximal}, TRPO \cite{schulman2015trust}, A2C, and A3C \cite{mnih2016asynchronous} directly optimize the policy rather than finding it through a value function. \cite{hafner2023mastering} introduced a general reinforcement learning algorithm named DreamerV3. DreamerV3 improves its behavior by learning an environment model and imagining future scenarios, achieving cross-domain learning.

\section{Preliminary}

\subsection{Causal Structural Model}
Structural causal models (SCMs) are commonly represented using causal graphs \cite{pearl2018book,cai2024granger}. A causal graph is comprised of a set of causal variables, denoted as $\mathbf{V}$, which serve as the nodes, and a set of causal edges, represented as $\mathbf{E}$, which connect these variables, forming the edges of the graph. This can be formally expressed as $\mathcal{G}(\mathbf{V}, \mathbf{E})$. A causal edge from variable $v_i$ to variable $v_j$ indicates that $v_i$ is a direct causal factor for $v_j$. Within our framework, we conceptualize objects in the environment and player-related information as nodes within the causal graph.  

\subsection{Causal Intervention}
\begin{figure}[ht]
    \centering
    \includegraphics[scale=0.5]{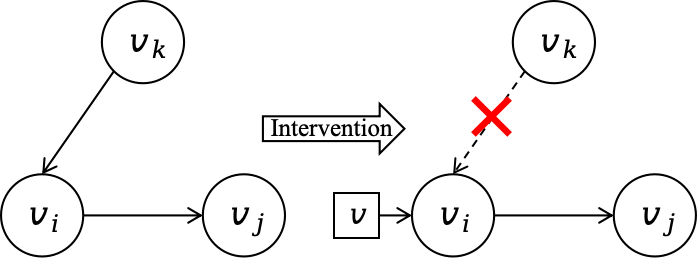} 
    \caption{Causal intervention}
    \label{intervention}
\end{figure}

In causal inference, an \textit{intervention} involves applying a change to a system or process to see how this change affects other variables or outcomes, defined as $P(v_j \mid do(v_i=v))$, where $do()$ denotes the do-operator \cite{pearl2009causality,pearl2018book}, which represents forcing $v_i=v$ and observing the resulting changes in $v_j$. As shown in Figure \ref{intervention}, before the intervention, the causal effect propagates through $v_k \to v_i \to v_j$. By intervening on the variable $v_i$, all incoming edges to $v_i$ are removed, and the causal effect flows solely through $v_i \to v_j$. Intervention helps us understand and confirm causal relationships, not just correlations. By examining the environmental feedback of interventions, we can better understand how variables are related, which helps us evaluate and predict the real influence of actions on a system, aiding both research and decision-making.

\section{Problem Formulation}
In this work, we aim to leverage causality to enhance the ability of LLM to understand the environment for effective decision-making. Given the inherent uncertainty and partial observability in real-world environments, we utilize a Partially Observable Markov Process (POMDP) \cite{sondik1971optimal}, which characterizes the interaction dynamics between an agent and the environment. Formally, a POMDP is described as a 7-tuple $(S, A, P, \Omega, O, \gamma, R)$. Here, $s \in S$ represents the state of the environment, $a \in A$ denotes an action sampled from a goal-conditioned policy $\pi$ \cite{liugoal}. The state transition probability $P(s' \mid s, a)$ specifies the likelihood of transitioning to a new state $s'$ after taking action $a$ in the current state $s$. The observation $o \in \Omega$ is determined by the observation function $O(o \mid s', a)$, which provides the probability of $o$ given the next state $s'$ and the action $a$. The reward function $R(s, a)$ quantifies the reward attained from executing action $a$ in state $s$, thereby reflecting the agent's benefit from specific actions. The discount factor $\gamma$ modulates the significance of future rewards in present decision-making, where a higher $\gamma$ value indicates a greater emphasis on long-term rewards.

Given the above setup and definitions, we propose a novel approach, causal-aware LLMs. Specifically, it first leverages the LLM to infer a causal graph $\mathcal{G}$ . This causal graph is continuously refined through iterative causal interventions (detailed in Section 4.3). The dynamically updated causal knowledge is then leveraged to generate a $goal$, which guides the RL agent in policy optimization. Therefore, our primary objective is to design a policy $\pi(a \mid o, goal, \mathcal{G})$ that incorporates the $goal$ derived from the causal graph $\mathcal{G}$, which maximizes the cumulative reward. This causal-aware mechanism enables the agent to make informed decisions by incorporating causal relationships within the environment.

\begin{algorithm}[t]
\caption{Causal-aware LLMs Framework}
\label{algo:framework}
\begin{algorithmic}[1] 
    \REQUIRE Observations $o$
    \ENSURE Policy $\pi$
    \FOR{$i$ in $train\_epoch$}
        \STATE \textbf{Learning Stage:} Learning a causal graph of the current environment information using a LLM from $o$.
        \STATE \textbf{Adapting Stage:} Updating causal graph $\mathcal{G}$ using causal interventions.
    \STATE \textbf{Acting Stage:} Using the updated $\mathcal{G}$ to assist in guiding the RL agent’s policy learning.
    \ENDFOR
\end{algorithmic}
\end{algorithm}

\section{Method}
\subsection{Framework}
In this section, we provide a causal-aware LLMs framework that leverages causal knowledge to facilitate policy learning in the interaction between the LLM and the RL agent, which enhances the reasoning and adaptive ability of LLM.
As illustrated in the Figure \ref{framework}, our proposed framework contains three stages: 1) \textit{Learning} for obtaining causal knowledge; 2) \textit{Adapting} for updating causal knowledge; and 3) \textit{Acting} for causal-aware decision making. 

In detail, in the Learning Stage, for the initial environment, we first provide the textual information of the environment to the LLM, which learns the causal variables and an initial causal graph between those variables. In this stage, the causal graph may contain some inaccurate causal relationship due to LLM's hallucinations. In the Adapting Stage, to update the inaccurate causal relationship, we make use of the idea of causal intervention to adapt. If an agent acts to change the state of a variable $s^i$, then the states of the other variables change, which indicates that $s^i$ is the ancestor of those other variables, This idea is incorporated into LLM to let it update the causal graph according to the textual data. In the Acting Stage, based on the learned causal knowledge, another LLM will generate a sub-goal according to the overall achievements the agent needs to accomplish. Driven by this sub-goal, the agent will do a series of actions for learning policy to obtain a reward in the environment. This causes a state transition and may update the information of the environment. The new environment information will be regarded as the input to LLM and the procedure is entered into the latest iteration. The procedure of our framework is summarized in Algorithm \ref{algo:framework}.

\subsection{Learning: Obtaining Causal Knowledge from LLMs}
To obtain causal relationships, we begin by extracting causal relations from the environment, which are subsequently used to construct a causal graph $\mathcal{G}$.
In the context of a dynamic system, we collect the observations $o_t$ over a series of processes from the environment, which can be converted into textual representations. We employ in-context learning with few-shot prompting to leverage the LLM’s pre-trained knowledge for extracting causal relations: $relations = LLM(o_t, \text{prompt})$. 
Based on the learned causal relations $s^i_t\rightarrow s^j_t$ $(i,j=1,2,\dots,n\ i \neq j)$, a causal graph $\mathcal{G}$ is constructed. Formally, the causal graph can be represented as a causal matrix $\mathcal{M}$. Specifically, the matrix element $\mathcal{M}[i][j]=1$ indicates the presence of a causal relationship between the causal variables $s^i_t$ and $s^j_t$, while $\mathcal{M}[i][j]=0$ indicates its absence. Note that in this stage, the causal knowledge may be incomplete or incorrect, due to the limited reasoning ability of the pre-training framework of LLM. And the incorrect causal knowledge will be refined in the following updating stage.
Detailed information on the prompt design is provided in Appendix A.9.

\begin{figure}[t]
    \centering
    \includegraphics[width=1\linewidth]{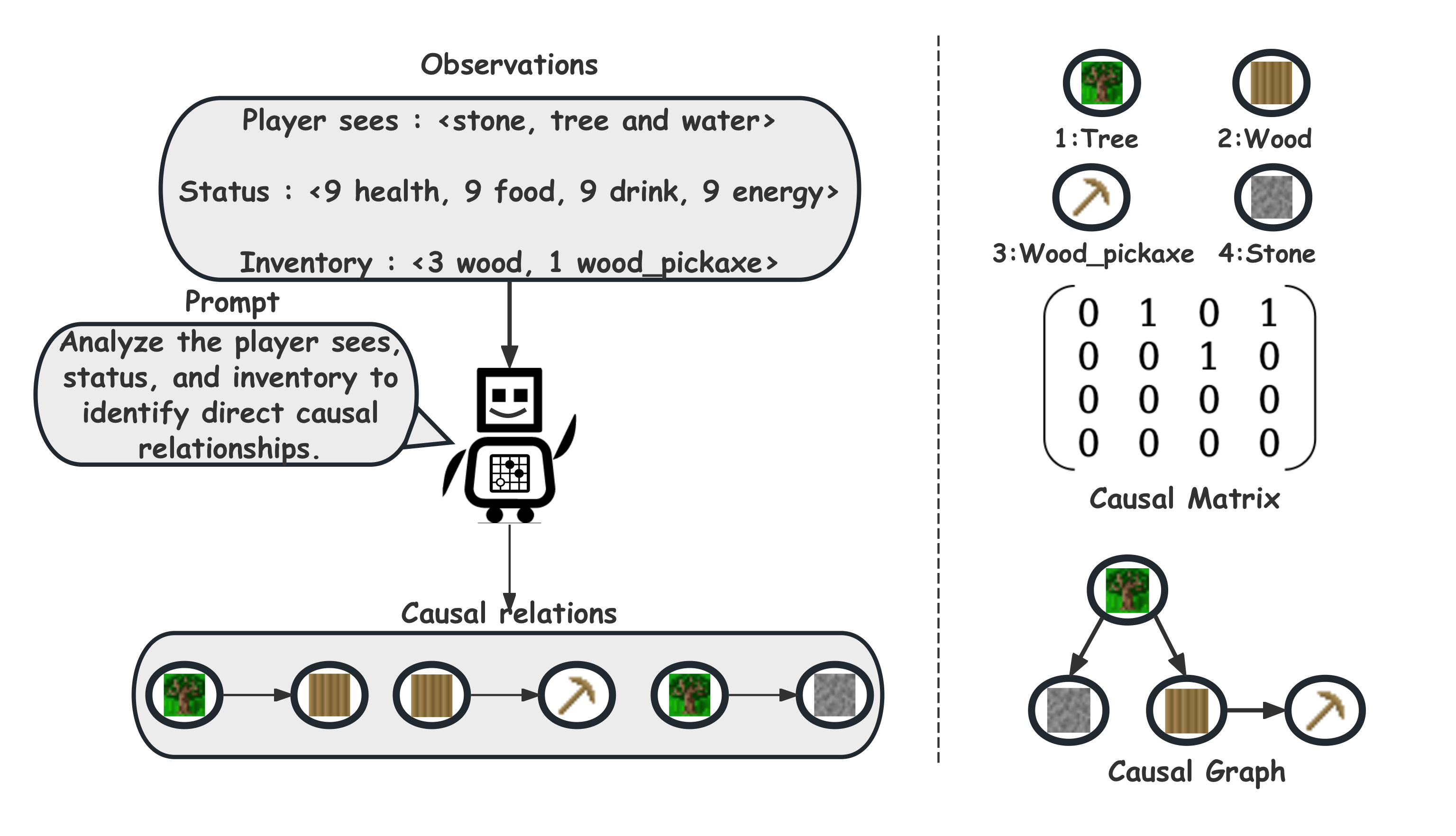}
    \caption{An example of learning causal knowledge from LLM.}
    \label{causal graph construction}
\end{figure}

\textbf{Example.} An example of the learning stage procedure is illustrated in Figure \ref{causal graph construction}. Based on current observation, LLM first extracts the causal relations including \( \text{$tree$} \rightarrow \text{$wood$} \), \( \text{$wood$} \rightarrow \text{$wood\_pickaxe$} \), \( \text{$tree$} \rightarrow \text{$stone$} \). These relations are then used to construct a causal graph.

\subsection{Adapting: Updating Causal Knowledge from Environment Feedback}

To update the causal knowledge, we first verify the correctness of the learned causal relations through causal intervention \cite{pearl2009causality} and environmental feedback, followed by the relationship updating process. 

To facilitate causal relationship verification, we design a valid environment $\texttt{Valid\_Env}$ through a three-step process. First, we construct an independent verification environment separate from the primary training environment. Second, we configure this environment to specifically support the agent's verification of causal relationships. Third, we disable model parameter updates during the verification process, ensuring that the agent evaluates causal relationships using only its currently learned policy. 

Subsequently, within this environment, we prompt the LLM to generate an appropriate goal based on the unverified relations and use this goal to sample actions $a_t$. We apply the action $a_t$ to the causal variable $v^i_t$ as an intervention and observe the changes in the effect variable $v^j_{t+1}$ at the subsequent time step $t+1$. After executing the action $a_t$ on $v^i_t$ (denoted as $do(v^i_t)$), we evaluate whether $p(v^j_{t+1}\mid do(v^i_t)) = p(v^j_{t+1} \mid v^i_t)$ holds. If it holds, it indicates that there is no causal relationship between those two variables; otherwise, a causal relationship exists. Through this approach, we can verify the correctness of unverified causal relationships. This mechanism enables the LLM to dynamically update the causal graph based on environmental feedback, ensuring the accuracy of its causal understanding.

\textbf{Example.} In the Crafter environment, to update the causal relation between causal variables \text{$tree$} and \text{$wood$}, we first generate a valid environment. In the generated environment, we will place a \text{$tree$} at each of the four corners of the eight squares surrounding the agent. Then, a well-trained agent will perform an action: $chop\ tree$. Finally, we check the status of \text{$wood$}. In such a case, its value will change, which means that the causal relation between \text{$tree$} and \text{$wood$} is correct. According to the result, we will update the causal knowledge as \( \text{$tree$} \rightarrow \text{$wood$} \).

\subsection{Acting: Causal-aware Decision Making}

To enable causal-aware exploration and decision-making, we leverage the causal knowledge acquired through the previous two stages and propose a goal-guided acting framework that integrates this knowledge with LLMs. The framework incorporates causal relationships into the decision-making process to guide both goal generation and policy learning.

First, inspired by goal-conditioned policy learning, we decompose complex tasks into multiple sub-goals while respecting the causal relations encoded in $\mathcal{G}$. This decomposition establishes a hierarchical structure that supports gradual task completion while maintaining causal consistency. The goals can be represented in various forms, such as ``reaching a specific location", ``collecting a designated item" or ``completing a specific sequence of tasks". As illustrated in the Acting Stage in Figure \ref{framework}, we utilize LLM to generate appropriate goal: $goal = LLM(prompt, \mathcal{G})$, where the generation is explicitly guided by the causal knowledge captured in $\mathcal{G}$.

Subsequently, the agent executes actions $a_t$ through a causal-aware policy: $a_t \sim\pi(a_t \mid s_t, goal, \mathcal{G})$, where the policy $\pi$ is trained to maximize rewards while considering causal constraints. To address the challenge of sparse rewards during early training stages, we introduce a reward enhancement mechanism that leverages semantic alignment. Specifically, we augment the standard reward function with an additional reward term, computed as the cosine similarity between the embeddings of the current goal and environmental observations. This semantic alignment reward provides dense feedback signals that facilitate more efficient exploration and decision-making based on the underlying causal model.

\textbf{Example.} Consider the Crafter environment as an example. After updating the causal relations between \text{$tree$} and \text{$wood$} as \( \text{$tree$} \rightarrow \text{$wood$} \), if the goal generated by LLM is ``obtaining wood", the agent's policy recognizes that wood acquisition causally depends on the presence of trees. Consequently, the action is sampled as either $chop\ tree$ or $find\ tree$. When the agent executes these actions under the causal constraint (i.e., first finding a tree, then chopping it), it receives positive rewards, which reflects that it makes an efficient decision. 

\section{Experiment}
\subsection{Environment Setting}
We use \textit{Crafter} environment \cite{hafner2022crafter} and Meta-Llama-3-8B-Instruct as our based LLM to evaluate the performance of our framework. All experiments are conducted on an NVIDIA GeForce 4090 with 24G \footnotemark.

\footnotetext{The code is available at: \url{https://github.com/DMIRLAB-Group/Causal-aware_LLMs}}

\paragraph{Environment Details.} Crafter is a 2D version of Minecraft, with a game map size of 64×64 and a player’s field of view of 9×7. The environment contains various interactive objects including natural resources (e.g., trees, iron blocks), hostile entities (e.g., zombies), and craftable items. The primary objective for players is to collect and create various items categorized on an achievement tree, aiming to unlock as many achievements as possible. The partial observability of the environment, coupled with the need for long-term planning and resource management, makes Crafter particularly suitable for evaluating our framework's causal reasoning capabilities. Further details about Crafter can be found in the Appendix A.8.  

\paragraph{Evaluation Metrics.}
To evaluate the effectiveness of our proposed framework, we assess the agent's capabilities through its performance in the Crafter environment. Specifically, we focus on two key metrics: success rate, and overall score. 
The success rate measures the frequency with which the agent unlocks various achievements during training. The overall score reflects the agent's balanced performance across all tasks, computed using the geometric mean of individual achievement success rates:
\begin{equation}
    Score = \exp \left( \frac{1}{N} \sum_{i=1}^{N} \ln (1 + s_i) \right) - 1,
\end{equation} 
where $N$ is the total number of achievements that need to be unlocked. In the Crafter environment, $N=22$. We adopt the geometric mean rather than the arithmetic mean because it better reflects the agent's balanced performance across all achievements, being more sensitive to underperformance in any single task. These metrics collectively demonstrate not only the agent's proficiency in specific skills but also indicate our framework's effectiveness in facilitating causal understanding and strategic decision-making in complex environments.
\subsection{Baselines}
To provide a thorough evaluation context for our framework, we conduct comprehensive comparisons against three categories of current state-of-the-art methods:

\textbf{LLM-based Methods.} We compare our method with classical reinforcement learning methods that incorporate LLMs, including React \cite{yao2023react}, Reflexion \cite{shinn2023reflexion} and AdaRefiner \cite{zhang2024adarefiner}, which leverage the text generation and reasoning capabilities of LLMs to enhance RL agents. These comparisons aim to demonstrate that leveraging causal models for environmental understanding can enhance LLM's reasoning abilities and adaptability to novel environments.  

\textbf{RL-based Methods.}
We also compare our method with RL methods mentioned in the Crafter benchmark, including classic approaches such as PPO (Resnet version) \cite{moon2024discovering}, DreamerV2 \cite{hafnermastering}, DreamerV3 \cite{hafner2023mastering} and Rainbow \cite{hessel2018rainbow}. Both DreamerV2 and DreamerV3 try to understand the environments to enhance decision-making capability of agent. Rainbow integrates multiple improvement methods, which significantly enhances the algorithm's performance. These comparisons aim to validate that the integration of LLMs and causal models enables environmental understanding, thereby improving decision-making performance.

\begin{figure*}
    \centering
    \includegraphics[width=1\linewidth]{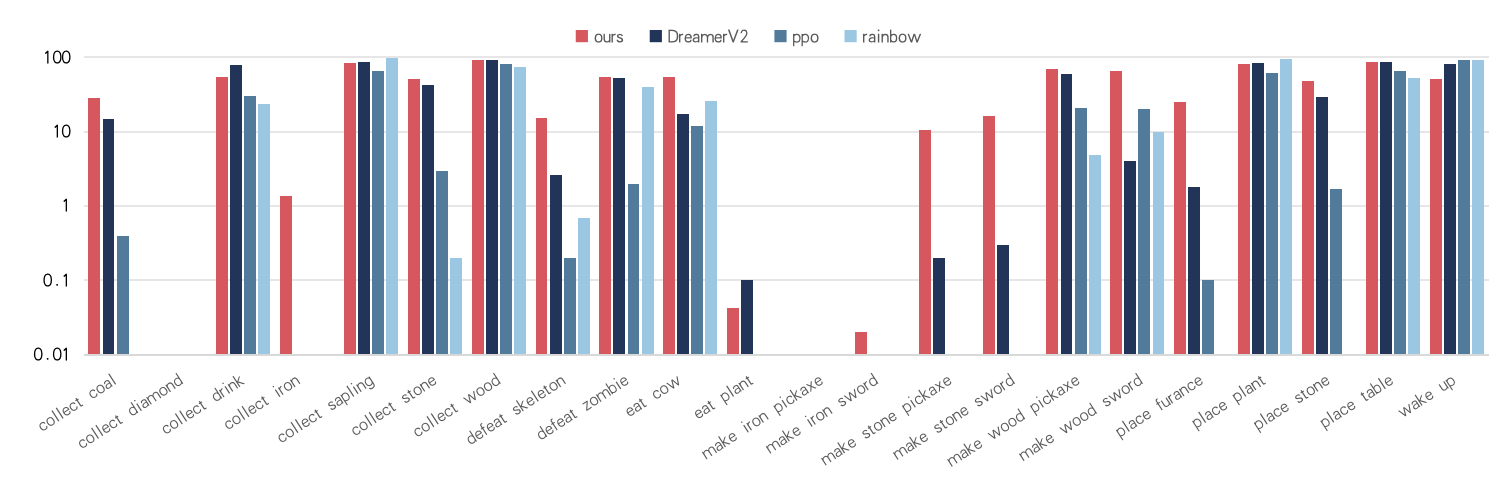}
    \caption{Success rates of obtaining 22 achievements in log scale.}
    \label{fig:achievements}
\end{figure*}

\textbf{Additional references.} These methods are based on additional references, including the results of human experts \cite{hafner2021benchmarking}, SPRING \cite{wu2024spring}, and random experiments. SPRING explores the use of games and puzzles to cultivate and study people’s reasoning abilities, including the reasoning processes in games like Crafter and Minecraft, as well as reasoning strategies in general puzzles and problem-solving games. These comparisons aim to validate the effectiveness of understanding environmental mechanisms in decision-making.

\subsection{Results and Analysis}
In the experimental results, for methods without publicly available implementations, we reference performance metrics from their original publications, maintaining consistency in experimental protocols and evaluation frameworks wherever possible. Table \ref{tab:results_differentMethods} and Figure \ref{fig:achievements} demonstrate the performance of our methods and baseline methods. 

According to Table \ref{tab:results_differentMethods}, our method significantly outperforms all baseline methods except for human experts and SPRING. These results strongly indicate that incorporating causal modeling into causal-aware LLMs enables the agent to develop a more comprehensive understanding of the environment's underlying mechanics. This enhanced understanding not only accelerates policy learning but also significantly improves the agent's ability to handle complex scenarios. While SPRING shows competitive performance in the one-million-step evaluation, primarily due to its advantage of leveraging pre-defined game rules and environmental knowledge as prior information, our method demonstrates more compelling long-term benefits. Specifically, in the early training stages, our approach faces initial challenges in causal learning and graph updates due to insufficient training. However, the five-million-step results reveal a notable performance improvement, where our method surpasses SPRING. This superiority can be attributed to the agent's enhanced ability to perform the acting stage to interact with the environment, and the iterative adapting stage to refine the causal graph, which leads to a more accurate and deeper understanding of the environment's mechanism.

Regarding the success rates of obtaining different achievements, as illustrated in Figure \ref{fig:achievements}, our method shows significant improvements in unlocking and achieving deeper accomplishments as the number of training steps increases. Specifically, we observe that with a deeper understanding of the environment, the agent’s ability to achieve more challenging accomplishments also increases. This indicates that integrating environmental causal knowledge not only enhances the agent’s understanding of the environment but also improves the agent’s performance in complex environments.

\begin{table}[!t]
    \centering
    \small
    \begin{tabular}{lcc}
        \toprule
        Method Type & Method & Score ($\%$) \\
        \midrule
        \multirow{2}{*}{Ours}
         & Causal-aware LLMs (@1M) & \textbf{18.9 $\pm$ 0.53} \\
         & Causal-aware LLMs (@5M) & \textbf{33.6 $\pm$ 0.02} \\
        \midrule
        \multirow{4}{*}{RL-based}
         & Rainbow(@1M) & 4.3 $\pm$ 0.2 \\
         & DreamerV2(@1M) & 10.0 $\pm$ 1.2 \\
         & DreamerV3(@1M) & 14.77 $\pm$ 1.42 \\
         & PPO(Resnet)(@1M) & 15.6 $\pm$ 1.66 \\
        \midrule
        \multirow{4}{*}{LLM-based}
         & ReAct(GPT-4)(@1M) & 8.3 $\pm$ 1.2 \\
         & Reflexion(GPT-4)(@1M) & 11.7 $\pm$ 1.4 \\
         & AdaRefiner(@1M) & 15.8 $\pm$ 1.4 \\
         & AdaRefiner(@5M) & 28.2 $\pm$ 1.8 \\
        \midrule
        \multirow{3}{*}{Additional refs}
         & Human Experts & 50.5 $\pm$ 6.8 \\
         & SPRING(+prior)(@1M) & 27.3 $\pm$ 1.2 \\
         & Random(@1M) & 1.6 $\pm$ 0.0 \\
        \bottomrule
    \end{tabular}
    \caption{Scores (mean $\pm$ std) of our method and baselines on 22 Crafter tasks.}
    \label{tab:results_differentMethods}
\end{table}

\begin{table}[!t]
    \centering
    \begin{tabular}{l c}
        \toprule
        \textbf{Method (@1M)} & \textbf{Score (\%)} \\
        \midrule
        Causal-aware LLMs & \textbf{18.9} $\pm$ \textbf{0.53} \\
        \midrule
        Ours w/o learning & 14.89 $\pm$ 0.75 \\
        Ours w/o adapting & 14.67 $\pm$ 0.73 \\
        PPO (ResNet) & 15.6 $\pm$ 1.66 \\
        \bottomrule
    \end{tabular}
    \caption{Ablation study of causal-aware LLMs.}
    \label{tab:ablation}
\end{table}

\subsection{Ablation Study}
To evaluate the contribution of each stage in the causal-aware LLMs framework, we conducted comprehensive ablation studies. These experiments were specifically designed to assess the impact of different stages in our proposed framework, thereby providing insights into their combined effects on the overall performance.

As illustrated in Table \ref{tab:ablation}, when the causal graph modeling in the learning stage is absent, the framework’s decision-making process lacks the guidance of the causal graph, significantly reducing its effectiveness. In this scenario, the agent can only gradually adapt to the environment through RL policy learning. 

When causal interventions for refining the causal graph are absent, the results are even worse than those without the learning stage. This is because incorrect causal relationships not only fail to provide proper guidance for the decision-making process but also introduce noise, misleading the RL agent in its decision-making. 

The PPO (Restnet) method is employed to evaluate the contribution of integrating causal knowledge and LLMs into RL agents. Without the assistance of LLMs and causal reasoning mechanisms, PPO (Restnet) exhibits substantially inferior performance, particularly demonstrating notable deficiencies in decision-making and reasoning capabilities. These comparative experiments highlight the significant performance gains achieved through the integration of causal knowledge and LLM components into the RL framework.

Therefore, these results further demonstrate the importance of causal knowledge in understanding environmental mechanisms and enhancing decision-making capabilities, particularly in complex decision-making tasks.
\subsection{Convergence Analysis}

We analyzed the convergence of the Causal-aware LLMs during the training process. As shown in Table \ref{tab:different stage}, since the causal graph is not fully established in the early training stage, the RL agent’s policy learning in the acting stage relies mainly on reward signals. This stage accounts for only the first 10\% of the training cycle. As training progresses with the help of reward signals, the agent gradually acquires basic skills, allowing it to refine the causal graph accurately. At this point, the agent begins leveraging accurate causal relationships to guide policy learning, making its policy more accountable and effective in the acting stage.

\begin{figure}[t]
    \centering
    \includegraphics[width=1\linewidth]{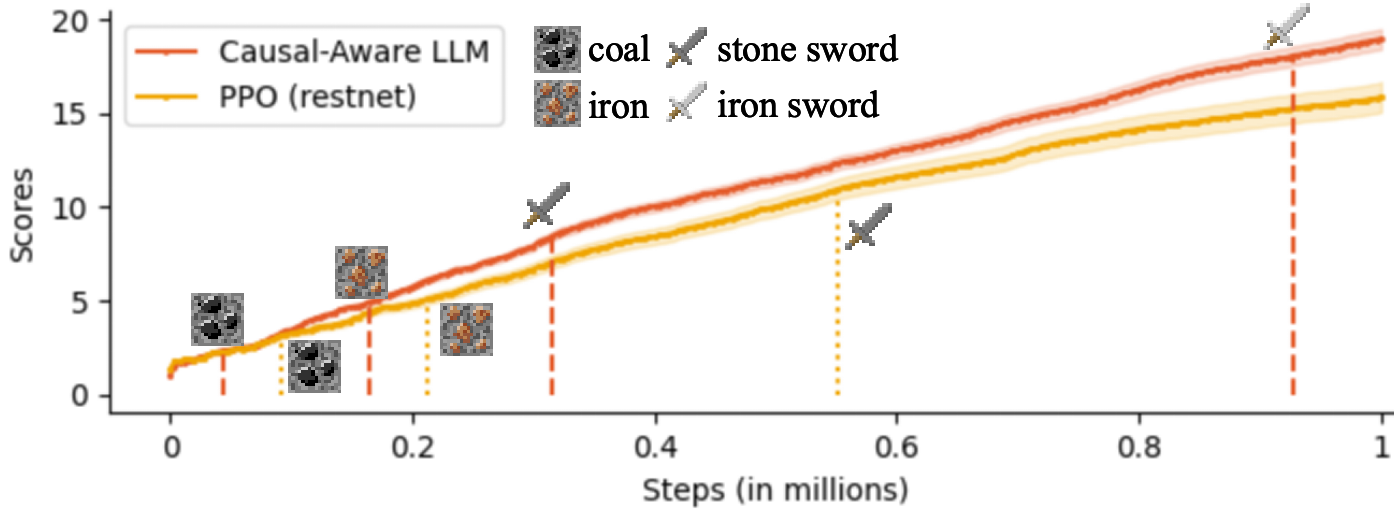}
    \caption{Comparative score of Causal-aware LLMs and PPO (ResNet) in achievement unlock times.}
    \label{fig:unlock}
\end{figure}

We also analyzed the unlock times for several achievements, as illustrated in Figure \ref{fig:unlock}. The results demonstrate that our proposed method significantly accelerates the achievement unlocking process, outperforming the PPO (ResNet) by approximately 0.2M training steps. While both approaches showed comparable performance during the initial 0.1M steps, the integrated causal knowledge in our method proved increasingly advantageous as training progressed. This advantage became particularly evident in complex scenarios, enabling the agent to master sophisticated achievements such as ``$make\ iron\ pickaxe$".

\begin{table}[ht]
    \centering
    \resizebox{\columnwidth}{!}{
        \begin{tabular}{ccccccc}
            \toprule
            Steps range (Million) & [0, 0.1] & [0.1, 0.2] & [0.2, 0.4] & [0.4, 0.6] & [0.6, 0.8] & [0.8, 1] \\
            \midrule
            Causal-aware LLMs & 2.35 & 3.47 & 7.97 & 11.4 & 14.56 & 17.71 \\
            PPO (ResNet) & 2.34 & 3.18 & 6.69 & 10.02 & 12.85 & 15.01 \\
            \midrule
            Gap & +0.01 & +0.29 & +1.28 & +1.38 & +1.71 & +2.70 \\
            \bottomrule
        \end{tabular}
    }
    \caption{Performance across different training stages.}
    \label{tab:different stage}
\end{table}

\section{Conclusion}
In this paper, we propose a Causal-aware LLMs framework to enhance decision-making, by introducing the structural causal model (SCM) into LLMs. The causal-aware LLMs framework is based on the ``learning-adapting-acting" manner. 
These three stages iterate continuously until convergence is achieved. Experimental results in the Crafter environment demonstrate that by utilizing causal knowledge, the agent achieves substantial improvements in task execution and environmental adaptation, outperforming traditional baseline methods. 
This implies the possibility that our approach can be applied to complex scenarios especially where the causal mechanism can be modeled. 
In early training, limited agent capability hinders accurate causal knowledge updates from environmental feedback. While LLM or domain priors help, performance remains poor without them. Future work could dynamically adjust the adaptation phase based on the agent’s learning progress.

\section*{Ethical Statement}
There are no ethical issues.

\section*{Acknowledgments}

This research was supported in part by the National Science and Technology Major Project of China (2021ZD0111502), the National Science Fund for Excellent Young Scholars (62122022), National Natural Science Foundation of China (U24A20233, 62206064, 62406078), Guangdong Basic and Applied Basic Research Foundation (2025A1515010172, 2023B1515120020), the Guangzhou Basic and Applied Basic Research Foundation (2024A04J4384). 

\bibliographystyle{named}
\bibliography{ijcai25}

\begin{thebibliography}{}

\bibitem[\protect\citeauthoryear{Ashwani \bgroup \em et al.\egroup }{2024}]{ashwani2024cause}
Swagata Ashwani, Kshiteesh Hegde, Nishith~Reddy Mannuru, Dushyant~Singh Sengar, Mayank Jindal, Krishna Chaitanya~Rao Kathala, Dishant Banga, Vinija Jain, and Aman Chadha.
\newblock Cause and effect: Can large language models truly understand causality?
\newblock In {\em Proceedings of the AAAI Symposium Series}, volume~4, pages 2--9, 2024.

\bibitem[\protect\citeauthoryear{Baumli \bgroup \em et al.\egroup }{2023}]{baumlivision}
Kate Baumli, Satinder Singh, Feryal Behbahani, Harris Chan, Gheorghe Comanici, Sebastian Flennerhag, Maxime Gazeau, Kristian Holsheimer, Dan Horgan, Michael Laskin, et~al.
\newblock Vision-language models as a source of rewards.
\newblock In {\em Second Agent Learning in Open-Endedness Workshop}, 2023.

\bibitem[\protect\citeauthoryear{Brown \bgroup \em et al.\egroup }{2020}]{brown2020language}
Tom Brown, Benjamin Mann, Nick Ryder, Melanie Subbiah, Jared~D Kaplan, Prafulla Dhariwal, Arvind Neelakantan, Pranav Shyam, Girish Sastry, Amanda Askell, et~al.
\newblock Language models are few-shot learners.
\newblock {\em Advances in neural information processing systems}, 33:1877--1901, 2020.

\bibitem[\protect\citeauthoryear{Cai \bgroup \em et al.\egroup }{2024}]{cai2024granger}
Ruichu Cai, Yunjin Wu, Xiaokai Huang, Wei Chen, Tom~ZJ Fu, and Zhifeng Hao.
\newblock Granger causal representation learning for groups of time series.
\newblock {\em Science China Information Sciences}, 67(5):152103, 2024.

\bibitem[\protect\citeauthoryear{Chen \bgroup \em et al.\egroup }{2024}]{chen-etal-2024-clear}
Sirui Chen, Mengying Xu, Kun Wang, Xingyu Zeng, Rui Zhao, Shengjie Zhao, and Chaochao Lu.
\newblock {CLEAR}: Can language models really understand causal graphs?
\newblock In Yaser Al-Onaizan, Mohit Bansal, and Yun-Nung Chen, editors, {\em Findings of the Association for Computational Linguistics: EMNLP 2024}, pages 6247--6265, Miami, Florida, USA, November 2024. Association for Computational Linguistics.

\bibitem[\protect\citeauthoryear{Du \bgroup \em et al.\egroup }{2023}]{du2023guiding}
Yuqing Du, Olivia Watkins, Zihan Wang, C{\'e}dric Colas, Trevor Darrell, Pieter Abbeel, Abhishek Gupta, and Jacob Andreas.
\newblock Guiding pretraining in reinforcement learning with large language models.
\newblock In {\em International Conference on Machine Learning}, pages 8657--8677. PMLR, 2023.

\bibitem[\protect\citeauthoryear{Gao \bgroup \em et al.\egroup }{2023}]{gao-etal-2023-chatgpt}
Jinglong Gao, Xiao Ding, Bing Qin, and Ting Liu.
\newblock Is {C}hat{GPT} a good causal reasoner? a comprehensive evaluation.
\newblock In Houda Bouamor, Juan Pino, and Kalika Bali, editors, {\em Findings of the Association for Computational Linguistics: EMNLP 2023}, pages 11111--11126, Singapore, December 2023. Association for Computational Linguistics.

\bibitem[\protect\citeauthoryear{Hafner \bgroup \em et al.\egroup }{2020}]{hafnermastering}
Danijar Hafner, Timothy~P Lillicrap, Mohammad Norouzi, and Jimmy Ba.
\newblock Mastering atari with discrete world models.
\newblock In {\em International Conference on Learning Representations}, 2020.

\bibitem[\protect\citeauthoryear{Hafner \bgroup \em et al.\egroup }{2023}]{hafner2023mastering}
Danijar Hafner, Jurgis Pasukonis, Jimmy Ba, and Timothy Lillicrap.
\newblock Mastering diverse domains through world models.
\newblock {\em arXiv preprint arXiv:2301.04104}, 2023.

\bibitem[\protect\citeauthoryear{Hafner}{2021}]{hafner2021benchmarking}
Danijar Hafner.
\newblock Benchmarking the spectrum of agent capabilities.
\newblock In {\em Deep RL Workshop NeurIPS 2021}, 2021.

\bibitem[\protect\citeauthoryear{Hafner}{2022}]{hafner2022crafter}
Danijar Hafner.
\newblock Benchmarking the spectrum of agent capabilities.
\newblock In {\em International Conference on Learning Representations}, 2022.

\bibitem[\protect\citeauthoryear{Hessel \bgroup \em et al.\egroup }{2018}]{hessel2018rainbow}
Matteo Hessel, Joseph Modayil, Hado Van~Hasselt, Tom Schaul, Georg Ostrovski, Will Dabney, Dan Horgan, Bilal Piot, Mohammad Azar, and David Silver.
\newblock Rainbow: Combining improvements in deep reinforcement learning.
\newblock In {\em Proceedings of the AAAI conference on artificial intelligence}, volume~32, 2018.

\bibitem[\protect\citeauthoryear{Kahneman}{2011}]{kahneman2011thinking}
Daniel Kahneman.
\newblock Thinking, fast and slow.
\newblock {\em Farrar, Straus and Giroux}, 2011.

\bibitem[\protect\citeauthoryear{Kiran \bgroup \em et al.\egroup }{2021}]{kiran2021deep}
B~Ravi Kiran, Ibrahim Sobh, Victor Talpaert, Patrick Mannion, Ahmad~A Al~Sallab, Senthil Yogamani, and Patrick P{\'e}rez.
\newblock Deep reinforcement learning for autonomous driving: A survey.
\newblock {\em IEEE Transactions on Intelligent Transportation Systems}, 23(6):4909--4926, 2021.

\bibitem[\protect\citeauthoryear{Kober \bgroup \em et al.\egroup }{2013}]{kober2013reinforcement}
Jens Kober, J~Andrew Bagnell, and Jan Peters.
\newblock Reinforcement learning in robotics: A survey.
\newblock {\em The International Journal of Robotics Research}, 32(11):1238--1274, 2013.

\bibitem[\protect\citeauthoryear{Lillicrap \bgroup \em et al.\egroup }{2015}]{lillicrap2015continuous}
Timothy~P Lillicrap, Jonathan~J Hunt, Alexander Pritzel, Nicolas Heess, Tom Erez, Yuval Tassa, David Silver, and Daan Wierstra.
\newblock Continuous control with deep reinforcement learning.
\newblock {\em arXiv preprint arXiv:1509.02971}, 2015.

\bibitem[\protect\citeauthoryear{Liu \bgroup \em et al.\egroup }{2022}]{liugoal}
Minghuan Liu, Menghui Zhu, and Weinan Zhang.
\newblock Goal-conditioned reinforcement learning: Problems and solutions.
\newblock In Lud~De Raedt, editor, {\em Proceedings of the Thirty-First International Joint Conference on Artificial Intelligence, {IJCAI-22}}, pages 5502--5511, 7 2022.
\newblock Survey Track.

\bibitem[\protect\citeauthoryear{Mnih \bgroup \em et al.\egroup }{2013}]{mnih2013playing}
Volodymyr Mnih, Koray Kavukcuoglu, David Silver, Alex Graves, Ioannis Antonoglou, Daan Wierstra, and Martin Riedmiller.
\newblock Playing atari with deep reinforcement learning.
\newblock {\em arXiv preprint arXiv:1312.5602}, 2013.

\bibitem[\protect\citeauthoryear{Mnih \bgroup \em et al.\egroup }{2016}]{mnih2016asynchronous}
Volodymyr Mnih, Adria~Puigdomenech Badia, Mehdi Mirza, Alex Graves, Timothy Lillicrap, Tim Harley, David Silver, and Koray Kavukcuoglu.
\newblock Asynchronous methods for deep reinforcement learning.
\newblock In {\em International conference on machine learning}, pages 1928--1937. PMLR, 2016.

\bibitem[\protect\citeauthoryear{Moon \bgroup \em et al.\egroup }{2024}]{moon2024discovering}
Seungyong Moon, Junyoung Yeom, Bumsoo Park, and Hyun~Oh Song.
\newblock Discovering hierarchical achievements in reinforcement learning via contrastive learning.
\newblock {\em Advances in Neural Information Processing Systems}, 36, 2024.

\bibitem[\protect\citeauthoryear{Pearl and Mackenzie}{2018}]{pearl2018book}
Judea Pearl and Dana Mackenzie.
\newblock {\em The book of why: the new science of cause and effect}.
\newblock Basic books, 2018.

\bibitem[\protect\citeauthoryear{Pearl}{2009}]{pearl2009causality}
Judea Pearl.
\newblock {\em Causality}.
\newblock Cambridge university press, 2009.

\bibitem[\protect\citeauthoryear{Schulman \bgroup \em et al.\egroup }{2015}]{schulman2015trust}
John Schulman, Sergey Levine, Pieter Abbeel, Michael Jordan, and Philipp Moritz.
\newblock Trust region policy optimization.
\newblock In {\em International conference on machine learning}, pages 1889--1897. PMLR, 2015.

\bibitem[\protect\citeauthoryear{Schulman \bgroup \em et al.\egroup }{2017}]{schulman2017proximal}
John Schulman, Filip Wolski, Prafulla Dhariwal, Alec Radford, and Oleg Klimov.
\newblock Proximal policy optimization algorithms.
\newblock {\em arXiv preprint arXiv:1707.06347}, 2017.

\bibitem[\protect\citeauthoryear{Shao \bgroup \em et al.\egroup }{2019}]{shao2019survey}
Kun Shao, Zhentao Tang, Yuanheng Zhu, Nannan Li, and Dongbin Zhao.
\newblock A survey of deep reinforcement learning in video games.
\newblock {\em arXiv preprint arXiv:1912.10944}, 2019.

\bibitem[\protect\citeauthoryear{Shinn \bgroup \em et al.\egroup }{2023}]{shinn2023reflexion}
Noah Shinn, Beck Labash, and Ashwin Gopinath.
\newblock Reflexion: an autonomous agent with dynamic memory and self-reflection.
\newblock {\em arXiv preprint arXiv:2303.11366}, 2(5):9, 2023.

\bibitem[\protect\citeauthoryear{Sondik}{1971}]{sondik1971optimal}
Edward~Jay Sondik.
\newblock {\em The optimal control of partially observable Markov processes}.
\newblock Stanford University, 1971.

\bibitem[\protect\citeauthoryear{Sun \bgroup \em et al.\egroup }{2023}]{sun2023adaplanner}
Haotian Sun, Yuchen Zhuang, Lingkai Kong, Bo~Dai, and Chao Zhang.
\newblock Adaplanner: adaptive planning from feedback with language models.
\newblock In {\em Proceedings of the 37th International Conference on Neural Information Processing Systems}, pages 58202--58245, 2023.

\bibitem[\protect\citeauthoryear{Touvron \bgroup \em et al.\egroup }{2023}]{touvron2023llama}
Hugo Touvron, Thibaut Lavril, Gautier Izacard, Xavier Martinet, Marie-Anne Lachaux, Timoth{\'e}e Lacroix, Baptiste Rozi{\`e}re, Naman Goyal, Eric Hambro, Faisal Azhar, et~al.
\newblock Llama: Open and efficient foundation language models.
\newblock {\em arXiv preprint arXiv:2302.13971}, 2023.

\bibitem[\protect\citeauthoryear{Watkins}{1989}]{watkins1989learning}
Christopher John Cornish~Hellaby Watkins.
\newblock Learning from delayed rewards.
\newblock 1989.

\bibitem[\protect\citeauthoryear{Wu \bgroup \em et al.\egroup }{2024}]{wu2024spring}
Yue Wu, So~Yeon Min, Shrimai Prabhumoye, Yonatan Bisk, Russ~R Salakhutdinov, Amos Azaria, Tom~M Mitchell, and Yuanzhi Li.
\newblock Spring: Studying papers and reasoning to play games.
\newblock {\em Advances in Neural Information Processing Systems}, 36, 2024.

\bibitem[\protect\citeauthoryear{Yao \bgroup \em et al.\egroup }{2023}]{yao2023react}
Shunyu Yao, Jeffrey Zhao, Dian Yu, Nan Du, Izhak Shafran, Karthik Narasimhan, and Yuan Cao.
\newblock React: Synergizing reasoning and acting in language models.
\newblock In {\em International Conference on Learning Representations (ICLR)}, 2023.

\bibitem[\protect\citeauthoryear{Zhang and Lu}{2024}]{zhang2024adarefiner}
Wanpeng Zhang and Zongqing Lu.
\newblock Adarefiner: Refining decisions of language models with adaptive feedback.
\newblock In {\em Findings of the Association for Computational Linguistics: NAACL 2024}, pages 782--799, 2024.

\bibitem[\protect\citeauthoryear{Zhang \bgroup \em et al.\egroup }{2024}]{zhang2024causal}
Yuzhe Zhang, Yipeng Zhang, Yidong Gan, Lina Yao, and Chen Wang.
\newblock Causal graph discovery with retrieval-augmented generation based large language models.
\newblock {\em arXiv preprint arXiv:2402.15301}, 2024.

\end{thebibliography}

\end{document}